# 5

# Modelling Immunological Memory

Simon Garrett[1], Martin Robbins[1], Joanne Walker[1], William Wilson[2], and Uwe Aickelin[2]

[1] Computational Biology Group, Department of Computer Science, University of Wales, Aberystwyth, SY23 3PG. Wales, UK. {smg,mjr00}@aber.ac.uk
[2] School of Computer Science (ASAP), University of Nottingham, Nottingham, NG8 1BB. England, UK. {w.wilson,uwe.aickelin}@notts.ac.uk

**Summary.** Accurate immunological models offer the possibility of performing high-throughput experiments *in silico* that can predict, or at least suggest, *in vivo* phenomena. In this chapter, we compare various models of immunological memory. We first validate an experimental immunological simulator, developed by the authors, by simulating several theories of immunological memory with known results. We then use the same system to evaluate the predicted effects of a theory of immunological memory. The resulting model has not been explored before in artificial immune systems research, and we compare the simulated *in silico* output with *in vivo* measurements. Although the theory appears valid, we suggest that there are a common set of reasons why immunological memory models are a useful support tool; not conclusive in themselves.

## 5.1 Introduction

One of the fundamental features of the natural immune system (NIS) is its ability to maintain a memory of previous infections, so that in future it can respond more quickly to similar infections [Sawyer 1931]. The mechanisms for immunological memory are still poorly understood and, as a result, are usually highly simplified during the construction of artificial immune systems (AIS).

Although all AIS are *inspired* by the immune system, see Chapter 3 of this book, here we study more detailed immunological *models*. Immune system models will be required by theoretical immunologists if there is to be a significant increase in the generation of new ideas in the field because computational simulation is considerably faster than laboratory experiments. So far, however, this has not been practical because the granularity of the simulations has been far too large, and single systems are able to either generate high-level, global immune simulations, or detailed but partial simulations, but not both.

We differentiate between a model and a metaphor. In AIS there are several metaphors, such as clonal selection methods, negative selection methods, and network methods that provide computational tools for the AIS practitioner. These are not models. Models are an attempt to create an artificial system that displays the same behaviours as another (normally natural) system. Metaphors simply use the natural system as inspiration for an algorithmic device.

Here we focus on the creation and use of immunological models in immunology. There may be side-effect benefits from these models that inspire the discovery of new computational methods in AIS, but that is not our central aim here. We outline a system, still under development, that can provide fast, detailed immune simulations, and which is beginning to suggest *in vivo* effects with enough accuracy to be useful as an immunology support tool. We choose immunological memory as our application area. This chapter:

- Provides a survey of immunological memory, including well-known theories, and a new immunological memory theory that may be of interest to AIS practitioners.
- Provides a survey of existing immune simulation systems.
- Describes how we built and tested a simple set of immunological memory models, and then expanded this approach to a more advanced, generic simulator.
- Describes how we tested the validity of a new theory of immunological memory [Bernasconi *et al.* 2002]. First we used the advanced immune simulator to generate *in silico* results from the new theory. Then, since this theory was generated in response to *in vivo* results, we evaluated the reliability of that theory by comparing our *in silico* results with the *in vivo* results.

Our advanced simulator is fast, even when simulating $10^8$ lymphocytes, the number present in a mouse. It also has the ability to simulate cytokine concentrations, which proved vital in simulating the work of [Bernasconi *et al.* 2002]. The simulator's speed and flexibility allows it to be applied to tasks that were previously impossible. Furthermore, our new simulator is not just a one-off immune simulation for a single task, rather it is designed from the ground-up as a reusable, flexible tool for research.

## 5.2 Background

### 5.2.1 Immune Memory

As with many aspects of immunology, our understanding of the processes underlying immunological memory is far from complete. As Zinkernagel et al say, in their seminal paper on viral immunological memory, *"Browsing through textbooks and* *authoritative texts quickly reveals that the definition of immunological memory is not straightforward."* [Zinkernagel et al. 1996]. Many of the questions they raised are still relevant almost ten years later. There are several theories, some of which appear mutually exclusive, and there is experimental evidence used to support almost all of these theories. Before examining the techniques for modelling theories of immunological memory, we need to discuss the theories themselves.

It is now widely accepted that hyper-sensitive *memory cells* exist, and research has been conducted in order to describe their attributes and behaviours, e.g. [McHeyzer-Williams & McHeyzer-Williams 2005]. Memory cells come in at least two varieties: memory B-cells and memory T-cells. These cells are formed during (or soon after) an immune response. Acute viral infections induce two types of long-term memory: humoral immunity, in which B-cells produce antibodies to tag cells infected by viruses, and cellular immunity, in which T-cells, activated by specific viral antigens, kill the virus-infected cells and also produce cytokines that prevent the growth of viruses and make cells resistant to viral infection[3].

It has been established that a memory of an infection is retained for several years or even decades [Sawyer 1931, Paul et al. 1951]. One way to measure the strength of this immune memory is by counting the population of specific memory cells. This figure tends to fall rapidly immediately after an infection, reaching a stable (but reproducing) level that is maintained over many years or decades, even in the absence of re-exposure to the antigen. The challenge facing immunologists is to discover how these cells are maintained.

Underlying these issues, it seems likely that some sort of homeostasis mechanism maintains a stable total population size of memory cells. Evidence suggests that the total number of memory cells in the body must remain roughly constant, and it has been shown that any increase rapidly returns to this resting concentration [Tanchot & Rocha 1995]; indeed, it is common sense that the number of cells could not increase indefinitely within the fixed volume of the immune system's host. One possible explanation for this is that memory cells (particularly T-cells) release cytokines that have an inhibitory effect on any enlarged antibody sub-population.

Overall, what differs in the theories of immunological memory is: (i) how memory cells are formed, and whether they are qualitatively different to other B- and T-cells, and (ii) how memory cells are maintained in the long term, so that the memory of the primary response is not lost by cell death.

**Long-Lived Memory Cell Theory:** Given that lymphocytes (both B- and T-cells) differentiate into 'memory cells', and that these memory cells are then highly responsive to the original antigenic trigger, the simplest way of implementing this in nature might be to invoke very long-lived memory cells. In this case, we would assume that there is no cell-division, the memory cells just live a very long time: moreover they *must* do so if they are to preserve immunity for many years. Is this possible, since the majority of our cells have a life-span much shorter than that of the body as a whole, and so cells are continually dying, and being renewed?

---

[3] from http://www.emory.edu/EMORY_REPORT/erarchive/2000/February/ er-february.21/2 _21_00memory.html.

Zinkernagel et al say that there's no convincing evidence for this type of phenotype [Zinkernagel *et al.* 1996] and current opinion, such as McHeyzer-Williams and McHeyzer-Williams', agree [McHeyzer-Williams & McHeyzer-Williams 2005]. Furthermore, experimental evidence contradicts the long-lived memory cell theory. A series of experiments on mice showed that memory T-cells can continue to divide long after any primary response [Tough & Sprent 1994, Tough *et al.* 1996]. Since a stable population is maintained, this means that memory cells must also be dying at a similar rate, and are therefore not as long-lived as originally believed.

Furthermore, it has been known for decades [Sawyer 1931, Paul *et al.* 1951] that antibody produced in response to an antigen can persist at significant levels in serum for years after the initial infection has occurred. Antibodies cannot survive in the body for a particularly long length of time, so we can conclude that plasma cells are sustaining these concentrations (the primary source of antibody). The problem is that plasma cells, in mice, have been shown to have a life-span of just a few months [Slifka *et al.* 1998], and that they are *only produced by differentiating memory cells*. This evidence shatters the theory of long-lived memory B-cells, and draws us to the conclusion that memory B-cells – like their T-cell equivalents – are being continually cycled long after any infection has been dealt with.

**Emergent Memory Theory**: To address these issues, a Emergent Memory theory suggests that there are no special memory cells as such, rather the effector cells naturally evolve towards highly specific cells, and are preserved from apoptotic death via some sort of 'preservase' enzyme, such as telomerase [Weng *et al.* 1997]. Although it is unlikely that emergent memory is stable in itself [Wilson & Garrett 2004], the process would explain how memory cells are created: they are just specialised forms of effector cells.

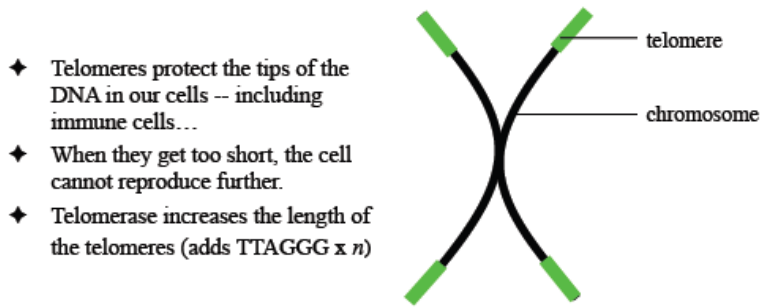

Fig. 5.1. Telomeres protect the tips of our chromosomes, and allow cells to reproduce successfully.

Each cell in our bodies can reproduce only a predefined number of times, as defined by the length of its *telomeres*. Telomeres are DNA sequences that 'cap' and protect the tips of our chromosomes, which are shorted each time the cell reproduces, indeed (Fig. 5.1), "... *each cycle of cell division results in a loss of 50 - 100 terminal nucleotides from the telomere end of each chromosome.*" [De Boer & Noest 1998]. What if the degree of telomere shortening were inversely proportional to the affinity between the cell's antibodies and antigen? In that case strongly matching immune cells would tend to survive longer than weakly matching ones.

This principle is not new in immunology – de Boer has suggested a model based on similar concepts [De Boer & Noest 1998]. Dutton, Bradley and Swain agree that the death rate is a vital component required in establishing robust memory. "*It stands to reason that activated cells must escape cell death if they are to go on to be memory. Thus, factors that promote the survival of otherwise death-susceptible T cells are candidates for memory factors*." [Dutton et al. 1998].

Consider the impact of this hypothesis in the context of different types of immune cells. Grayson et al state that, "... *memory T-cells are more resistant to apoptosis than naïve cells ... Re-exposure of memory cells to Ag* [antigen] *through viral infection resulted in a more rapid expansion and diminished contraction compared with those of naïve cells.*" [Grayson et al. 2002]. This indicates that memory cells would have lower (but not zero) death rates, and higher proliferation rates, so the the cell population would naturally contract to long-lived (i.e. high-affinity) cells over time.

Telomerase may not be the only biological mechanism that can explain the evolution of immune cells into longer lived, higher affinity memory cells, an alternative explanation underpinning the longer life-span of memory cells is provided by Zanetti [Zanetti & Croft 2001]: the "...*selection of B-cells destined to become memory cells takes place in GCs* [germinal centres] *and is controlled by the expression of intracytoplasmic molecules (Bcl-2 and Bcl-x) which prevent a form of cell death ... together with the concomitant suppression of signals from cell surface proteins that lead to death.*" Although differing from the telomerase hypothesis, the implications would be the same: memory cells appear to reflect normal immune cells that have naturally evolved to develop a lower death rate, ensuring their survival over other cells such as effectors.

The problem with the Emergent Memory theory is that it is very cell-specific. How can a concentration of cytokines ensure a high affinity cell lives longer than a lower affinity cell in almost the same location?

**Residual Antigen Theory:** Several reports suggest that protein antigen can be retained in the lymph node (e.g. [Perelson & Weisbuch 1997]), suggesting that normal lymphocyte function cannot remove *all* traces of a particular class of antigen. This is a natural result of the immune system being focussed on particular locations in the body. Whilst most antigenic material will be cleared by the immune system, causing an immune response, some antigenic material will escape a localised immune response long enough to reproduce. The immune system then quickly establishes a steady state between immune response and antigenic population size, and the immune system's population is stimulated by the normal hypermutation response.

Therefore, it is possible that the immune system does not completely remove all antigenic material from the host, either because small concentrations of antigenic cells may remain long enough to reproduce, or because the immune system itself has retained some of the antigenic material in follicular dendritic cells (FDCs). These FDCs then slowly release the antigenic material into the host, to stimulate a low-level immune response. Zanetti et al say, "*The prevailing view is that maintenance of B cell memory ... is a function of the persistence of antigen on FDCs ... only a few hundred picograms of antigen are retained in the long term on FDCs, but these small amounts are sufficient to sustain durable and efficient memory response.*" [Zanetti & Croft 2001]. In either case, this would keep the immune system active enough to sustain memory cell populations. This idea has been supported by research suggesting that B-cell memory is particularly sensitive to residual antigen [Tew *et al.* 1990].

In recent years however, compelling evidence has been presented suggesting that the cycling of memory T-cells continues to occur without any of the specific antigen being present [Lau *et al.* 1994], which would mean that these cells must be responding to some other stimulus. Although some debate has occurred [Manz *et al.* 2002, Zinkernagel 2002], this view is now widely accepted by immunologists [Antia *et al.* 2005].

An additional objection stems from an evaluation of the performance of such a system. How could it be efficient, from an evolutionary point of view, to expend resources on what is essentially a rote learning approach to memory? We know, from Machine Learning, that rote learning is the least efficient method of storing learned information, and it does not allow for generalisation. Although, there *is* an element of generalisation inherent in the Residual Antigen theory due to the memories of previous infections overlapping with new infections, and providing a (weak) generalised response, it is questionable whether there is enough generalisation to make this an effective source of immune memory.

It may seem that antigen persistence is important for a model of immune memory, to ensure that the high affinity memory cells are sustained over long periods, but there is another, related possibility. Perhaps memory cells do not need stimulation by antigen; they simply proliferate periodically. Would this represent another evolutionary step for an immune cell in order for it to differentiate into a memory cell? Grayson et al identified the discrepancy between the long term behaviour of memory cells and naïve cells and state that, "*... memory cells undergo a slow homeostatic proliferation, while naïve cells undergo little or no proliferation.*" [Grayson *et al.* 2002] (our emphasis). If this is the case, do memory cells actually need persistence of the antigen to survive?

Even if re-exposure is not necessary, Antia et al conclude that "*... estimates for the half-life of immune memory suggest that persistent antigen or repeated exposure to antigen may not be required for the maintenance of immune memory in short-* *lived vertebrates; however, ... repeated exposure may play an additional role in the maintenance of memory of long-lived vertebrates."* [Antia *et al.* 1998]. We choose to include antigen persistence in the model presented here.

**Immune Network Theory:** Network theory is based around the possibility that the immune system maintains and triggers memory by internal, not external stimulation. It suggests that immune cells, particularly lymphocytes, present regions of themselves that are antigenic to other immune cells. This causes cycles of stimulation and suppression, which, while begun by an external antigenic source, are continued and maintained even in their absence, and are thus a form of memory [Farmer *et al.* 1986]. A network of interactions between immune cells is widely believed to account for memory pool homeostasis [Zeng *et al.* 2005, Schluns & Lefrancois 2003], and certain immune cells are even able to form physically connected networks of tunneling nanotubules in vitro [Watkins & Salter 2005], but little evidence has been published recently in the major immunology journals for a strong role of the kind of co-stimulation described above.

**Heterologous and Polyclonal Memory Theories:** It has been observed that during an immune response, populations of memory T-cells unrelated to the antigen may also expand [Bernasconi *et al.* 2002, Tough *et al.* 1996], suggesting that perhaps serological memory could be heterologically maintained by a degree of *polyclonal stimulation* during all immune responses.

According to [Antia *et al.* 2005], two possible mechanisms have been suggested to explain these results - Bystander Stimulation and Cross-Reactive Stimulation:

(i) The Bystander Stimulation theory suggests that the antigen-specific T-cells produce a cytokine that stimulates all nearby (bystander) memory T-cells to divide. It has been suggested that bystander stimulation could be responsible for the continued cycling of memory B-cells, as well as for T-cells [Bernasconi *et al.* 2002]. The results of this high impact work showed that if memory B-cells are simultaneously exposed to an antigen that they are not specific to, and to the cytokine IL-15, they will undergo clonal expansion. This ability was shown to be unique to memory B-cells, and could not be repeated with their naïve equivalents.

(ii) the Cross-Reactive Stimulation theory is based on speculation that memory cells could be more sensitive to stimulation than naïve cells, and might therefore be stimulated by different antigens, perhaps even a self-antigen. In either case, it has been shown experimentally that memory T-cells specific to a particular antigen can be directly stimulated by a different, unrelated antigen [Selin *et al.* 1994].

Both of these theories suggest that once memory T-cells have been created, they can be stimulated during immune responses to unrelated antigen. The difference is that in one case the cells are directly stimulated by antigen, and in the other (polyclonal stimulation) they are stimulated by cytokines released by other, antigen-specific cells.

## 5.2.2 A Brief Survey of Immune Modelling

Mathematical Models: Mathematical models of immunological (sub)systems often use ordinary differential equations (ODE) or partial differential equations (PDE) to encapsulate their chosen dynamics (e.g. [Perelson 2002, Smith *et al.* 1999]). Perelson's HIV equations [Perelson 2002], and Smith's influenza dynamics [Smith *et al.* 1999], are illustrations of models of small parts of the immune system dynamics that have had significant benefits to human health, but which do not set out to model the immune system as a whole. In Chapter 4, we have already seen Perelson's detailed models of B cell and T cell receptors. When one considers the chemical complexity of amino acid binding it is not surprising that many balk at the idea of modelling the immune system at all. However, immunological simulations are possible because we observe gross-scale effects (such as primary/secondary responses) that are then modulated to a greater or lesser degree by small-scale processes, such as Perelson's discussion of B and T cell binding. Both are vital for truly accurate models, but larger scale models can be used successfully to explain gross-scale features of the immune system [Yates *et al.* 2001].

Immunological memory has also been modelled in a similar manner— the classic example being Farmer et al's work [Farmer *et al.* 1986] – but there are more recent attempts to model immunological memory too [Ahmed & Hashish 2003]. Although these models say a lot about certain details, they are not intended to be global models of immunological memory. For example, the important work of Antia et al on understanding $CD8^+$ T-cell memory [Antia *et al.* 2005] is based on a few, relatively simple equations. This is not to say that it is easy to generate such equations (it is not); rather, we are saying that the applicability of these equations is limited. Indeed, the difficulty in building and managing these equations is precisely the reason that a computational simulation approach is sometimes more appropriate.

Computational Models: Computational models are not as well established as mathematical models. Those that do exist are usually either population-based (entities that are tracked as they freely interact with each other), or cellular automata (entities that are tracked in a discrete grid-like structure, generally with local-only interactions [Wolfram 2002]). Nevertheless, computational models do have some advantages over mathematical models.

Firstly, it is possible to define, informally, the behaviour of a highly complex system, without formally defining it in terms of formal ODEs or PDEs—we can create a population of entities by mapping from objects in nature to objects in the computational simulation. Furthermore, many ODEs have no analytical solution and can *only* be solved by computational analysis, in software such as Matlab$^{TM}$ and Mathematica$^{TM}$.

Secondly, some forms of *in silico* experimentation may be difficult in mathematical models, and indeed in the immunology laboratory, such as tracking a single B-cell or antibody over its lifetime. It is possible, therefore, that computational immune simulators will provide the only means of investigating some immunological challenges.

In all computational simulations, we re-iterate the importance of the choice of binding mechanism, the type of cell-cell and cell-antigen interaction, (see [Garrett 2003] and Chapter 4 of this book), and we note that the few computational simulators that do exist are often underdeveloped and may not have been peer-reviewed by the academic community.

*ImmSim*: The work of Seiden et al, on ImmSim was the first real attempt to model the immune system as a whole, and it is still the only simulator to have been fairly widely peer reviewed [Kleinstein & Seiden 2000, Kleinstein *et al.* 2003]. It is similar in style to the work of Farmer et al [Farmer *et al.* 1986], but is a true simulation, not a set of ODEs[4].

*Simmune*: There are at least two "Simmune" immunology simulators: Meier-Schellersheim's version [Meier-Schellersheim & Mack 1999], which was developed in the late-1990s, and Smith and Perelson's version. Of the two, Meier-Schellersheim is the more advanced, implemented as a full cellular automata with the ability to define almost any rules that the user desired, whereas Smith and Perelson's was a relatively simple, unpublished Lisp simulation.

*Synthetic Immune System (SIS)*: Although SIS appears to be significantly faster and more powerful, it does much less. Simmune can simulate large numbers of complex interactions, whereas SIS is designed only to investigate self-nonself relationships. SIS is a cellular automata; it can only be found on the web[5].

*ImmunoSim*: Ubaydli and Rashbass's Immunosim set out to provide researchers with an "Immunological sandbox" - it was a customizable modelling environment that simulated cell types, receptors, ligands, cascades, effects, and cell cycles, with experiments run *in silico*. A key requirement was that it should have a purely visual interface, with no programming necessary. It received the Fulton Roberts Immunology prize (twice) from Cambridge University but does not appear to be available as a publication.

*Other systems:* These simulations [Castiglione *et al.* 2003, Jacob *et al.* 2004] are smaller scale than that proposed here, but have still had benefits to chemotherapy and immunology, and/or highlight problems that need to be overcome. Others have emphasised the importance of the binding mechanism, the type of cell-cell and cell-antigen interaction chosen, and the multitude of other possibilities that should be considered [Garrett 2003].

## 5.3 Basic Simulations

Our work with a set of Basic Simulations set out to explore the gross-scale behaviour of some of the theories just studied, while keeping the models as simple as possible – here, the only the interactions simulated are those between antibodies and antigen. This begs the question, "how simple can an effective model be?" Assuming Occam's razor applies, our answer is, "as simple as possible, and no simpler." However, the models described in this section are deliberately *too* simple. This is partly because no one knows how complex a simulation must be before it can accurately reproduce *in vivo* results, partly because by starting as simple as possible we get a lower limit on the computational performance of simple models, and partly (more importantly) because it lets us explore the dynamics underlying simple immune simulations, so that later additional complications can be viewed as modulations of this basic model. Note that the lack of complexity should not be seen as an indication that the models described in this section are trivial. Although simple, great care was taken to ensure they were as realistic as possible, as we hope will become clear.

---

[4] ImmSim currently to be found at http://www.cs.princeton.edu/immsim/software.html
[5] at: http://www.cig.salk.edu/papers/SIS_manual_wp_M.pdf

The Basic Simulations will also act as a primary validation for the *underlying mechanisms* of the more complex experiments. They do not validate any other aspect of the complex experiments. It is easier to verify and validate the performance of a simple model than a complex model; then if the complex and simple models share similar behaviour this partially validates the complex model. This raises another issue: how do we validate immunological models? If we apply standard Machine Learning methodology, where 'models' are 'hypotheses', then we should do some form of $k$-fold cross-validation to obtain a measure of the accuracy of the defined immunological hypotheses. But how do we do this when we have no well-established 'correct' data? To some extent, we can assume that if a model is able to *predict* what will be observed in nature, then the model is validated to some extent. Indeed, the ability to predict is one of the reasons for building models in the first place. We will return to this point later.

### 5.3.1 Basic Simulations: Methods and Materials

Each Basic Simulation was built from antibodies and antigen, and no models were allowed to directly create memory; memory had to evolve. This blurs the distinction between antibodies, B-cells and T-cells in order to explore the effects of immune cell/antibody proliferation in response to antigen. To indicate this blurring, we will call the simulated immune system elements 'reactive immune system elements', or RISEs. The RISEs were defined as being more likely to die as they got older; im- plemented by removing a RISE when $rnd().a > rnd().dr$, where $rnd()$ is a uniform random number generator, $a$ is the age of the RISE measured in generations from the current generation, and $dr$ is a death rate integer, which was set to 30. A constant
50 RISEs were added each generation. This led to a stable population size, which returned to the stable level despite external perturbations. Fig. 5.2 demonstrates this effect: despite a large influx of new RISEs (the large peak) and a small culling of RISEs (the small trough), stability is maintained. The size of the peaks were also reversed, with the same result that the population size returned to a stable level – note also the differences in scale between Fig.s 2(a) and 2(b), which show that the size of perturbation is irrelevant. This implements a simple homeostatic population of RISEs.

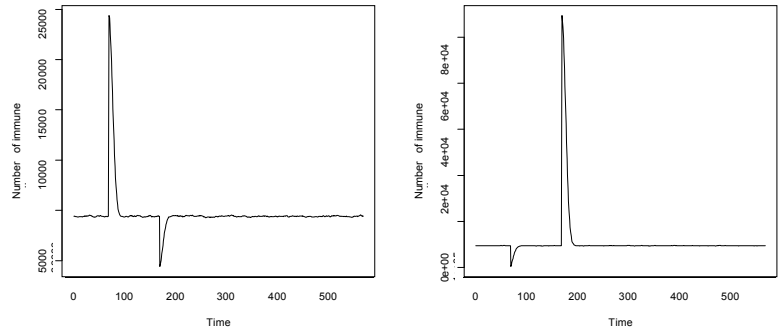

Fig. 5.2. The resting population of B-cells was in homeostasis. These graphs show the population stability that underlies all models that will follow. Any positive or negative change to the population size is quickly corrected, and stable population size is restored.

Antigen populations were 'injected' into the system as a whole, at a predefined times. The primary infection was always at generation 70, and the secondary infection was either at generation 120 ('smallGap' experiments), or generation 420 ('bigGap' experiments), to test the short- and long-term memory abilities of the population. An antigen was removed once is was bound to an RISE, and binding could only occur when the similarity between the RISE and antigen was within a distance of
100. The RISEs could take any value between zero and 10,000, and the antigen always had a randomly chosen value of 3.3, fixed at this value for all tests. In all cases we assume that the strongest affinity RISE will bind with the antigen. We implement this by a form of tournament selection, whereby the strongest matching RISE of ten randomly chosen RISEs is chosen to be the one that actually binds. Our more complex simulation, presented later in this chapter, uses simulated chemotaxis.

For each experiment, we measured the total number of RISEs, the total number of antigen, and the number of RISEs with affinity in the ranges, [0.01-0.1), [0.1-1), [1-10), [10-100), [100-1,000), [1,000-10,000) and [10,000-100,000). We recorded this information every generation for 600 generations.

### 5.3.2 Basic Simulations: Experiments and Tests

We performed the following experiments and tests.

**Memory By External Stimulation** These experiments tested the ability of the Basic Simulations to remember infections over a short and long period of time, assuming the only stimulation to be external, i.e. via antigenic interaction:

Control/None : On top of the homeostasis mechanism, we tested a standard implementation of clonal selection. This is activated by the presence of antigen, so that a good-matching RISE produces many clones, and a poor-matching RISE produced few clones. Furthermore, the good-matching clones are only slightly mutated from their parent cells, via a Gaussian centred on the parent, whereas the few poor-matching clones are often highly mutated, relative to their parents. This approximates Burnet's clonal selection theory [Burnet 1959] and acts as a control for these experiments. Since memory cells were not explicitly created, we would expect the RISE population to clear the antigen, and then forget the infection.

Emergent Memory : In the emergent memory tests, when a RISE was bound to antigen, the RISE's age was reduced in proportion to its affinity to the antigen, so that better-fitting RISEs tended to survive longer – this implemented the effects of 'preservase'. This should preserve the high matching RISEs to some extent, producing a form of memory.

Residual Antigen : Once the antigen population had been injected, a single antigen was then re-introduced into the simulation at random time intervals (on average, every three generations). Would this prevent the memory of the infection from being lost because this value is considerably smaller than $dr$? If so, under what conditions? It might be argued that this does not really represent residual antigen, as antigen are being reintroduced rather than maintained, however the purpose of this model is to show whether a small amount of stimulation can maintain memory, not to demonstrate mechanisms by which the antigen could be maintained, and thus in practical terms reintroduction performs the same role in our model as maintenance (keeping a small, stable population of antigen), with the advantage of allowing us to simplify the experiment.

Both Emergent Memory and Residual Antigen : Is there any benefit in implementing both the Emergent Memory and Residual Antigen theories?

**Memory By Internal Stimulation** These experiments tested the effects of adding internal stimulation to the Basic Simulations, so that one RISE could interact with another RISE, even in the absence of antigen. Although antibody-antibody interaction is not widely thought to be a form of memory in nature, it does occur, and is likely to have some function. These tests set out to suggest what that function might be. The graphs, described above, of affinity level distribution are of particular relevance to these experiments.

It is important to note that we do not use paratope-paratope binding here: i.e. we do not assume that a RISE/antibody's light chain will bind with the light chain of another RISE/antibody, for reasons outlined in [Garrett 2003] (e.g. the problems of positive feedback). Instead we shift ideal binding by 2,500 (in a circular range of 10,000) so that a RISE with value 1,000 would bind most strongly with another RISE of value 3,500. This means there would need to be a cycle of four RISEs if internal memory were to work. This implements paratope-epitope binding, although we note that there is still a functional relationship between the paratope's and the epitope's shape space, which is less than realistic, but was necessary to keep the simulation simple.

### 5.3.3 Basic Simulations: Results

**Memory By External Stimulation** The results are presented in Fig. 5.3 and Fig. 5.4. These are averages of ten runs. Remembering that the population is completely renewed on average every 30 generations, even the short gap experiments (left column of graphs; 50 generations between infections) should not have shown any memory of the previous infection.

In the "None" graphs (top line) we actually see a slight increase in response, but this is not statistically meaningful – there is no memory of previous infections. The statistics we used were the Wilcoxon Signed Rank Test, and the results are tabulated in Table 5.1. This test allowed us to decide when the difference in height between the primary and secondary responses was significant, and the ratio expresses the extent of that difference. This non-parametric test was chosen because it is likely that the secondary response is conditioned by the primary response, and that the data are *not* normally distributed.

The affinity graphs in Fig. 5.4 indicate there is an increase in RISEs that have affinities in the < 0.1, < 1.0 and < 10.0 ranges, but there is no memory between infections.

The "Emergent" results show a distinct secondary response in the short gap experiment, because the population members that were able to successfully bind were preserved beyond 30 generations; however, this effect is not enough to allow memory to persist over the big gap because the antibodies that were effective against the primary infection, tended to die over that time period. Nevertheless, the results indicate that memory can be preserved for at least 50 generations.

Now the affinity graphs show that high affinity RISEs are maintained between the infections that are separated by a small time gap, and how these types of RISE drain away over the longer time gap so that the simulated immune system needs to begin again to find a high affinity response to the antigen.

The "Residual Antigen" tests have a similar pattern in Fig. 5.3, with the population stimulated enough by the on-going, low-level antigen to promote a secondary response in the short gap experiment. In the big gap experiment, however, the effect is not statistically significant.

The affinity graphs show an elevated number of high- to mid-range affinity RISEs (in the < 1, < 10 and < 100 ranges) but indicates the very high affinity RISEs return to lower levels by 200 generations. This explains why the secondary response was not sufficient to be statistically meaningful when the antigenic injections were separated by a large gap.

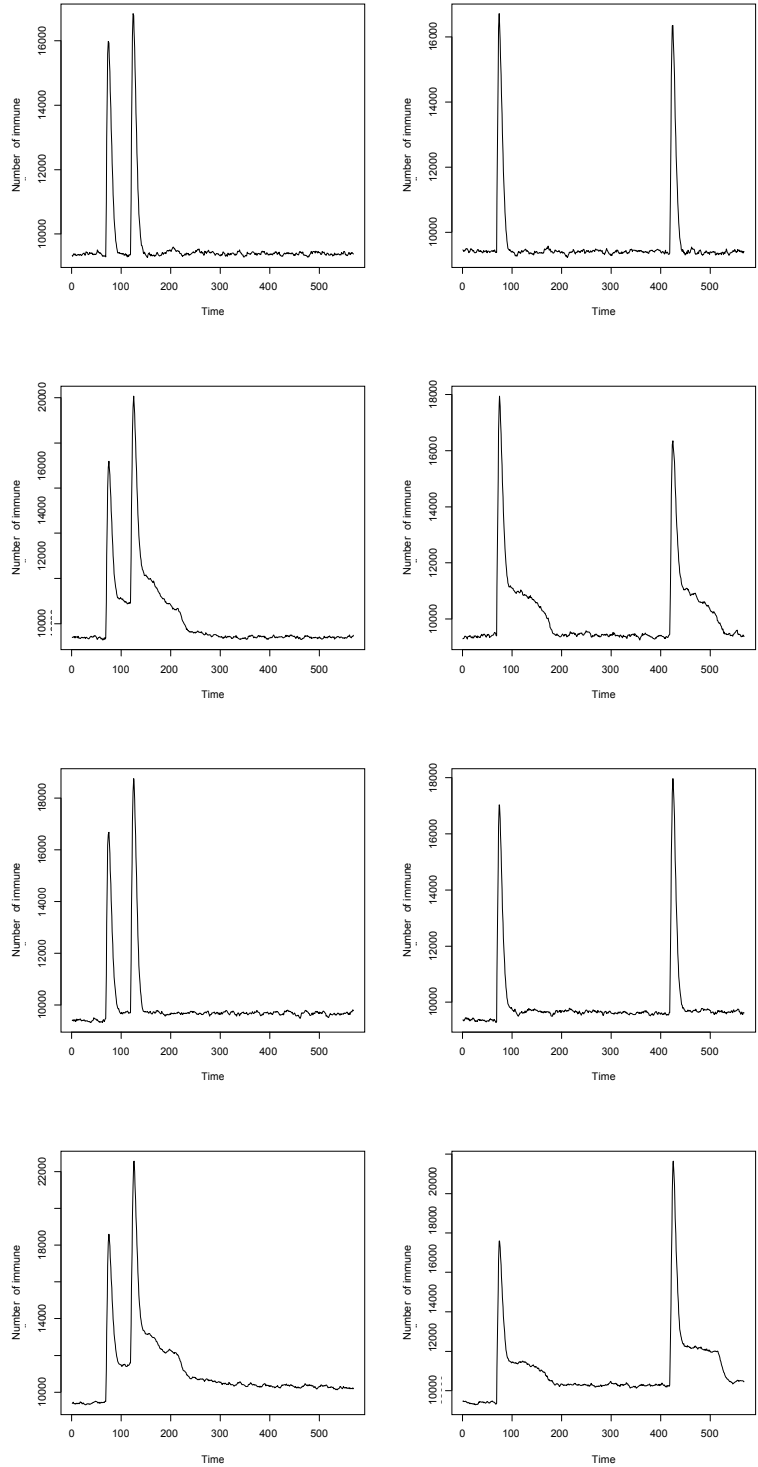

Fig. 5.3. Graphs of the theory simulations, "None", "Emergent", "Residual" and "Both" (top to bottom, in order) for a small time gap (50 generations, left column) and a longer time gap (350 generations, right column), averaged over 20 runs.

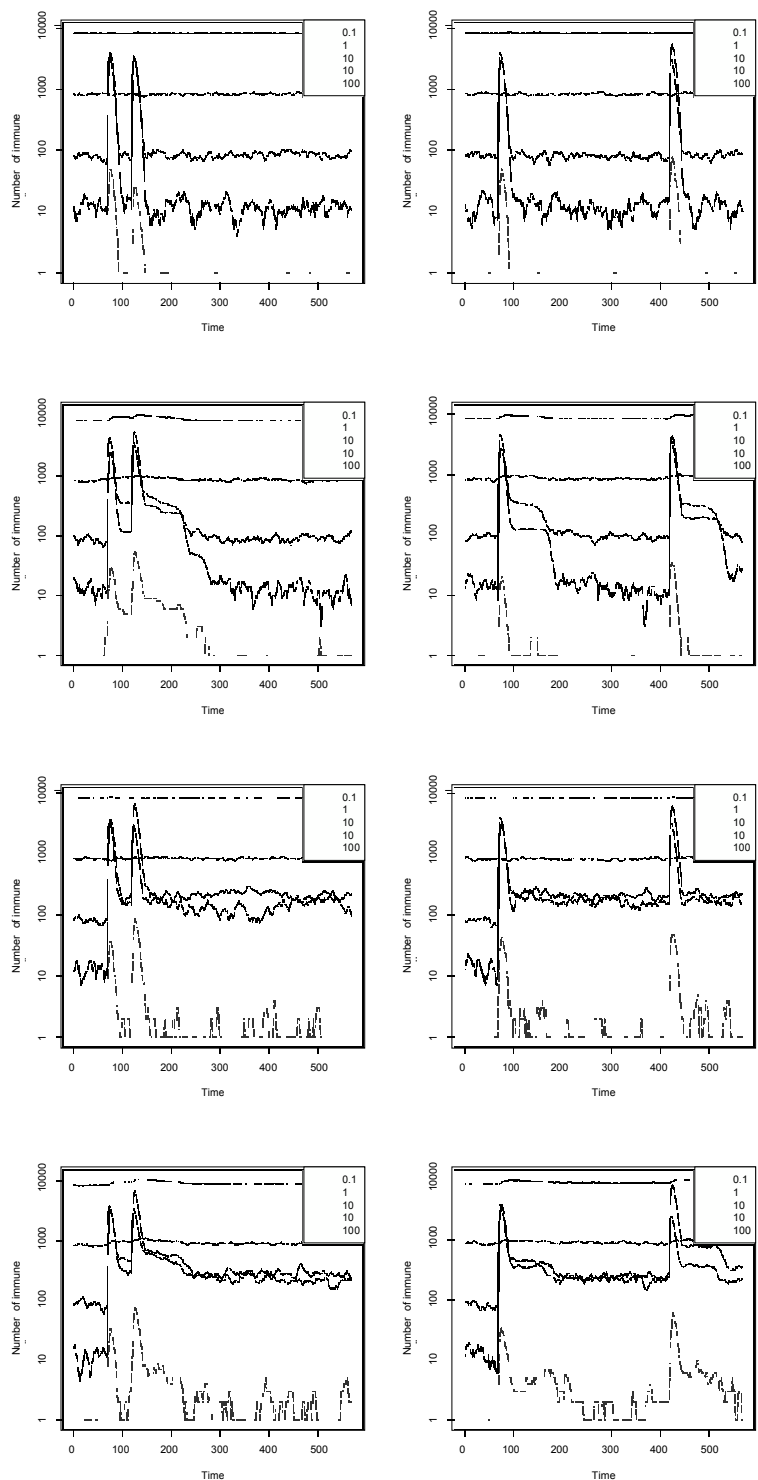

Fig. 5.4. Graphs of the theory simulations, "None", "Emergent", "Residual" and "Both" (top to bottom, in order) for a small time gap (50 generations, left column) and a longer time gap (350 generations, right column), averaged over 20 runs.

With "Both" emergent and residual antigen implemented, the story is different. Now, we see strong secondary responses for both short and big gaps, although there is a slight sustained, global increase in RISE population after the first infection.

The affinity graphs also show that the high affinity < 0.1 RISEs never returned to the zero mark. This appears to have been crucial in maintaining a powerful secondary response, and corresponds to the existence of high affinity memory cells in nature.

Some may ask why the residual antigen phenomenon does not explain immune memory on its own. If the amount of residual antigen were high enough then surely the immune response would be enough to remember that infection? Indeed, this is true, but at the cost of a permanently raised antibody population level, which is not seen in nature. At the extreme, if the infection were to persist at the same high levels then it is obvious that the memory would not be lost, because the infection would be continuous and on-going, but this is also not a realistic state of affairs, except in pathological cases, such as in elderly patients who are infected with cytomegolovirus [Perelson 2002]. The level chosen is one that only very slightly raises the antibody population size: it is enough to maintain memory over a short period, but not in the longer term.

Furthermore, Residual Antigen does not explain why better matching cells tend to survive and worse matching cells tend to die off; nor does it explain how memory cells can naturally emerge as a result of immune cell evolution. As a result, both the apoptosis reduction (or telomerase memory maintenance mechanism), and the re-stimulation mechanism are required to evolve an effective immune response.

**Memory By Internal Stimulation** The results are presented in Fig. 5.5 and Fig. 5.6. For each run, each RISE attempted to bind to the RISE with the highest affinity out of ten randomly chosen RISEs. There does not appear to have been any memory effect; indeed, the opposite seems true – as soon as any subpopulation increased in size out of proportion to the population as a whole, the network effect reduced the size of that subpopulation. This made the levels in Fig. 5.6 more *stable* than the comparative graphs in Fig. 5.4.

We conclude that the memory effects of immune networks are limited — at least the types of network that we have implemented here. Since our aims in these basic experiments are to produce simple models of immunological interactions, we use non-symmetric, paratope-epitope binding, in which knowing that A binds B does not imply B binds A. In contrast, AIS network algorithms tend to use paratope-paratope binding because it is of interest from a computational point of view, even if it is less biologically tenable.

| Experiment | p-Value | 99% | Ratio |
|---|---|---|---|
| None Small Gap | 0.240 | No | 0.948 |
| None Big Gap | 0.955 | No | 1.011 |
| Emergent Small Gap | 0.0000957 | Yes | 0.852 |
| Emergent Big Gap | 0.225 | No | 1.067 |
| Residual Small Gap | 0.00318 | Yes | 0.890 |
| Residual Big Gap | 0.332 | No | 0.945 |
| Both Small Gap | 0.0000957 | Yes | 0.822 |
| Both Big Gap | 0.0000957 | Yes | 0.813 |
| Network Small Gap | 0.765 | No | 0.999 |
| Network Big Gap | 0.896 | No | 1.001 |
| All Small Gap | 0.0000942 | Yes | 0.850 |
| All Big Gap | 0.000315 | Yes | 0.832 |

Table 5.1. Results of the Wilcoxon Signed Rank Test for difference between the size of the two peaks in each experiment. The p-values are shown to 3 significant figures and whether or not the difference can be regarded as significant at the 99% confidence level. The smaller the p-values the greater the degree of confidence that there is a difference between the primary and secondary responses, with 1.0 being zero confidence, and 0.0 being 100% confidence. The ratio gives the size and direction of the difference between the two peaks.

## 5.4 Experiments Using the Sentinel System

### 5.4.1 Method and Materials

The simulations that form the basis of this chapter were modelled using our soft- ware, 'Sentinel'. Sentinel is an agent-based complex system simulation platform for immunology and AIS research that currently exists as a prototype. Its design is based largely around the principles of cellular automata, with the environment di- vided into a discrete grid of locations. Entities within the simulation are free to move around in this environment, but are only able to respond to events that occur within closely neighbouring cells. 'Engines', such as those used in computer games for managing graphics, physics, etc., manage the physical and chemical interactions that occur within this environment.

The physics engine allows accurate simulation of the physical properties of agents, restricting their movements according to attributes such as size, mass or energy output. Whereas many simulations or differential equation models are exclusively based on cells that exhibit some form of Brownian motion, entities (cells) in Sentinel move according to the chemical stimuli they receive, their motor capabilities, and external forces acting upon them. The physics engine ensures that movement is as realistic as possible, and is a novel feature of our system.

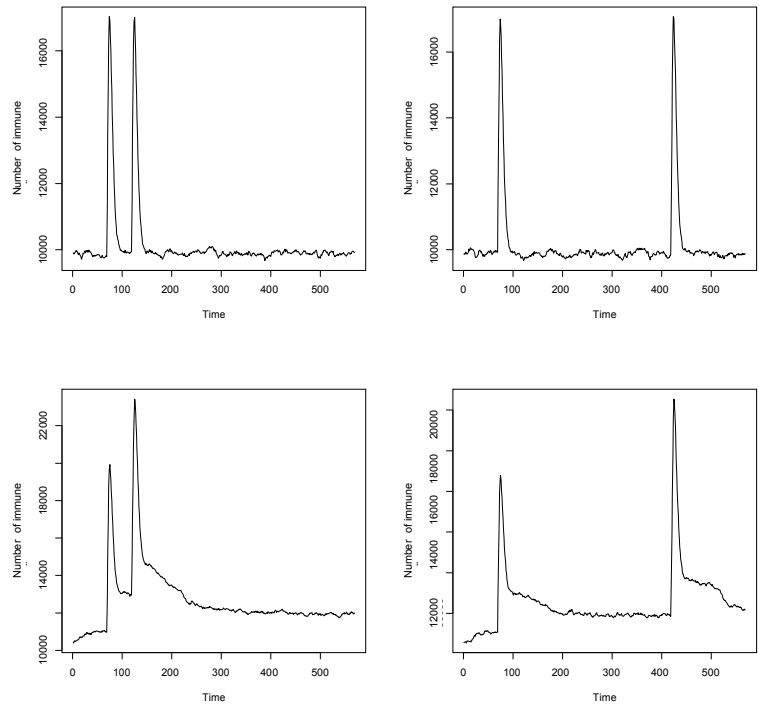

Fig. 5.5. Graphs of the theory simulations, "Network" and "All" (top to bottom) for a small time gap (50 generations, left column) and a longer time gap (350 generations, right column), averaged over 20 runs.

A chemistry engine is responsible for managing chemical and biochemical reactions, and also the distribution of extra-cellular molecules throughout the environment. For example, if a cell releases a particular kind of cytokine at its location, the chemistry engine will cause that cytokine to gradually disperse across the environment (see Fig. 5.7, right, for an example map of densities) by diffusion. This feature is essential for the accurate simulation of cell movement by chemotaxis – the process by which immune cells move towards higher concentrations of chemotactic factors, i.e. chemicals that attract them. It also enables a cell to influence a larger expanse of its environment than would typically be allowed in a cellular automata.

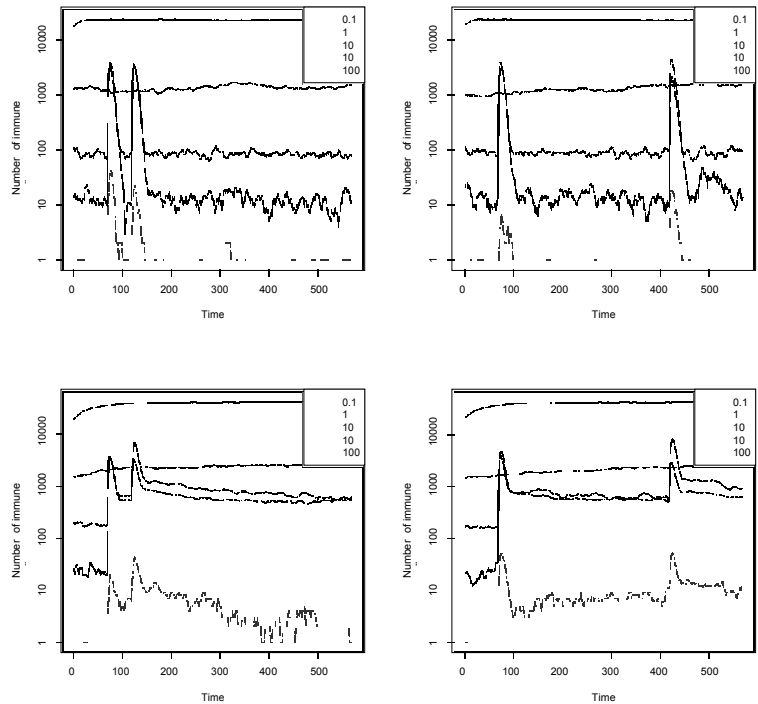

Fig. 5.6. Graphs of the theory simulations, "Network" and "All" (top to bottom) for a small time gap (50 generations, left column) and a longer time gap (350 generations, right column), averaged over 20 runs.

The implementation of chemotaxis is another novel feature of Sentinel. Cells *in vivo* are able to respond to various chemotactic molecules by detecting density gradients, and moving towards the highest or lowest density of that agent [Ramsay 1972]. The dispersal of chemotactic molecules in Sentinel is calculated by dispersing molecules from each location in the simulation to its neighbours over time. A cell in the simulation is able to access its eight neighbouring locations to find out the densities there, and retrieve the highest or lowest density of a particular molecule. It can then use this information to move accordingly.

Given a set of entities and chemicals (B-cells, antibodies, memory cells, cytokines, etc.), the influence of the physics and chemistry engines is defined by a number of *rules*. These rules define when an entity can interact with another cell, and the nature of that interaction; how one cell releases chemicals, or other entities, into its near environment, and any global features, such as blood flow that affect all entities and chemicals.

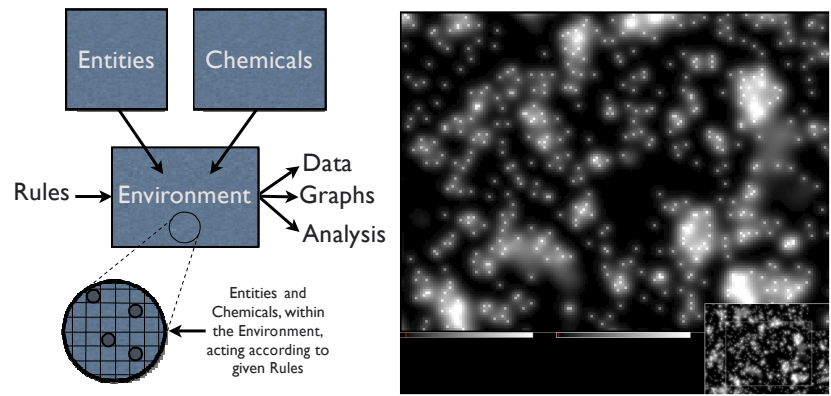

Fig. 5.7. (left) The structure of the Sentinel system. (right) Sentinel models the diffusion of chemicals to implement realistic chemotaxis and, crucially, to model the effects of cytokines (see text). The main figure shows the different concentrations of chemicals over a detailed view of the simulator's simulation environment. The inset shows the location of the detailed view in the whole space being modelled.

Having defined the simulation model, by choosing the entities, chemicals and rules, the simulator is run and information is output according to user-defined data-feeds. These data can then be viewed in the form of various graphs and samples, or streamed to log files for analysis, all within the Sentinel system. It seems likely that this simulator architecture will be useful in other areas too, such as biochemistry and abstract work in genetic and evolutionary computing.

The simulator is complemented by an Integrated Development Environment (IDE), that provides a set of powerful tools for the rapid development of new models. The drag-and-drop graphical interfaces allows the user to quickly choose sets of agents and establish the links between them, and to set up and connect areas of the en- vironment, and describe the rules of physics that will operate within them. A code editor allows users to develop Java-based extensions to these basic models, with the assistance of automated code-generation tools, and a comprehensive Application Programmers Interface (API) that provides general-purpose functions for manipu- lating agents and the environment. In many respects, the system is somewhat similar in nature to platforms such as Robocode[6], but far more powerful.

Sentinel can simulate several million cells, hundreds of millions of antibodies, and their interactions, on a typical high-end desktop. Although this figure varies depending upon the complexity of the model, Sentinel appears to be the most powerful simulator currently available, especially in view of the complex interactions that it is simulating. Sentinel's ability to simulate diffusion is very important – cytokine signalling between cells is a vital part of immunology. Indeed, one of the following experiments could not have been implemented without this ability.

---

[6] See http://robocode.sourceforge.net

### 5.4.2 Sentinel Experiments and Tests

**Sentinel Validation Tests**: Before using Sentinel to evaluate Bernasconi et al's theory, we validated its performance. Both the validation and the evaluation models ran with of the order of $10^8$ B-cells. We recapitulated the "None", "Emergent" and "Residual" experiments, as in the previous section, but did not implement "Network" because it had little value for our goals here. By implementing the same tests as the Basic Simulations, we set out to show that Sentinel would work at least as well as the Basic Simulations. If the results are qualitatively the same then we will have demonstrated that Sentinel can reproduce previous results. Each of our simulations were run ten times, in order to ensure that the results were consistently reproduced.

**Sentinel 'Theory Evaluation' Experiment**: This experiment is designed to explore the veracity of Polyclonal Activation Memory, via simulation – something which has not been done before. We could not use our Basic Simulation tool because the experiment required implementation of cytokine gradients (of IL-15), and needed to be performed on a much larger scale to obtain meaningful results. Only Sentinel could meet these requirements.

The construction of Bernasconi et al's model is based on the theory described in [Bernasconi *et al.* 2002]. They suggested their theories as a result of *in vivo* ex- periments, and claim that the experimental results provide compelling evidence for bystander stimulation of memory B-cell populations. The comprehensive set of re- sults published in [Bernasconi *et al.* 2002] will be tested against the data from our simulation, so our aim is to simulate the implications of Bernasconi et al's theory, and assess whether it could indeed be responsible for the *in vivo* results that they observed. Despite our validation efforts, the process described above is fairly limited and the process of parameterising any simulation is complex, therefore we can only safely look for qualitative similarities in between the results of Bernasconi et al and those produced by Sentinel.

### 5.4.3 Assumptions

In constructing these Sentinel models, a number of assumptions were made. These have been kept consistent through all the simulations conducted.

Repertoire: Sentinel's simulation repertoire included B-cells, antibodies, antigen, as well as a signalling chemical. It was more complex in terms of the entities used, and used many orders of magnitude more antibodies, than the RISEs in the Basic Simulations.

Longer-lived memory cells: Memory B-cells live longer than their naïve equivalents. In nature, a naïve B-cell tends to live for about 24 hours unless it receives stimulus, at which point it is "rescued, and may go on to live for a few months [Bernasconi *et al.* 2002] This is reflected in our models.

Antigen: Antigen does not reproduce or mutate during the simulation.

Simplified binding: As in the basic simulations, and in order to provide the best possible performance, a simplified binding mechanism was used. A strain of antigen is given a number between 0 and 20,000, which remains constant across the population. Every new B-cell is assigned a random number within that range, and the binding success is measured as the distance between the two numbers.

Clonal selection: In response to antigen, B-cells undergo clonal selection and hyper-mutation, as described by Burnet's 1959 theory. [Burnet 1959]. Cells that have been cloned retain the binding integer value (see previous bullet point) of their parent, mutated in inverse proportion to its binding strength.

Simplified Immune Repertoire: The simulation consists of B-cells, antibodies and antigen, plus one signalling cytokine. B-cell T-cell interaction is not simulated in these tests, but are planned (see Further Work). We needed to keep the model as similar to our previous system as possible (no plasma cells) to make the validation process as meaningful as possible.

### 5.4.4 Sentinel Results

### Sentinel's Validation Results

The results in Fig. 5.8 show that Sentinel correctly produces a secondary response to a repeat infection of the same antigen, for both memory theories. Furthermore, Sentinel's results show that the Residual Antigen model maintained a considerably higher population of memory cells and antibodies – down to only about $10^6$ anti- bodies before second infection, compared to of the order of $10^1$ for the other memory models. This relates to the Basic Simulations that showed the residual antigen pop- ulations had more antibodies.

In both simulators, the models of the Emergent, 'preserveron' theories sustained good short-term memory, and in both simulators we observed the memories stored in this manner failing when the cells carrying them died. Unless we accept that the primary immune response produces memory cells that live for years, such models will always result in an immune memory that fades over time.

The model of the Residual Antigen theory sustained a stable level of memory cells in both simulators, and was able to produce a substantial secondary response regardless of the length of time between the first infection and subsequent re-infection. It appears to be a viable model of immune memory; however, the requirements to sustain such a system seem unlikely to be met in nature because the immune system would have to produce such material over a highly extended period. Indeed this point was debated several years ago [Matzinger 1994a].

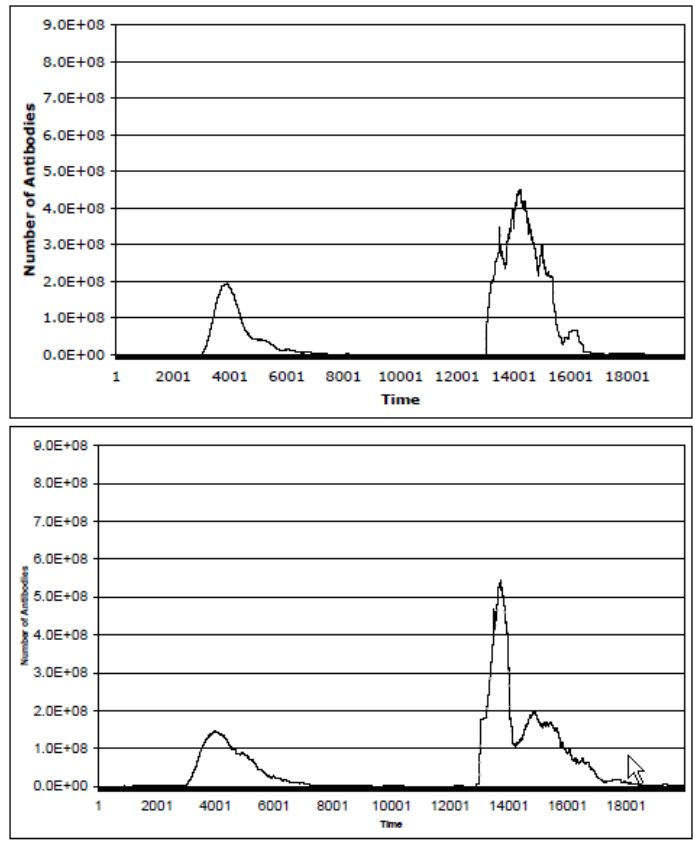

Fig. 5.8. Validation graphs for the number of cells over arbitrary time for: (i) the Emergent/'Preserveron' model, and (ii) the Residual Antigen model. Antigen A is injected at t=3000, and t=13,000.

Although there are some differences in the details, such as the more pronounced secondary peak in the secondary response, we consider the two simulators similar enough to proceed with the qualitative comparison of the *in vivo* and *in silico* results.

One advantage of Sentinel is that we can now distinguish between the secondary responses from the various theories: (i) the 'Preservon' model has a wide response, but it does not lead to as many antibodies being created, and the after-response is small, and (ii) the Residual Antigen model has a sharp, medium height secondary response, with a much extended, exponentially decreasing after-response.

Our previous experiments were too coarse-grained to provide results that had meaningful differences, and the curves they produced were an almost perfect exponential followed by a slower, almost perfect exponential decrease. Interestingly, there is a slight 'wobble' at the end of the exponential decrease, which is also consistent with the after response we see in the graphs of Fig. 5.8. These experiments show that we can reproduce the results of the Basic simulations, but with finer resolution.

### Sentinel's 'Theory Evaluation' Results

Since we stated in the 'Experiments and Tests' subsection that we have not validated the finer-grained elements of Sentinel's results, we will compare the results, in a qualitative way. Fig. 5.9 shows two plots from Sentinel – each for different model parameters – and a presentation of the graph from [Bernasconi *et al.* 2002].

Note that the Anti-A plot, caused by re-injected Antigen A, in (top) and (middle) has a shallower peak than the plot of Anti-TT in the bottom plot. The parameter values for the (top) graph yield poor results, but in (middle) are better, assuming we use the section of graph from time index one to five. The need to find good parameters is discussed in the Further Work section below. Both parameter choices result in some features of the Bernasconi *et al* plot but the *relative* increases seem to indicate that here is some degree of match between the simulated (middle) and *in vivo* (bottom) results.

Although not perfectly confirmed, a simulation of Bernasconi et al's theory has been shown to be qualitatively reasonable, relative to the *in vivo* measurements. But what causes the quantitative differences? The disparities may be due to: (i) *incorrect modelling* of the Bernasconi et al theory; (ii) *lack of detail* in the model; (iii) *incorrect parameterisation* of that model, and/or (iv) a fundamentally *faulty theory* underlying the model. The next step is to isolate the cause of disparity. The first and last of these points can be addressed by opening a dialogue with Bernasconi's group, but points (ii) and (iii) will require significant further work, as described below.

In conclusion, the simulated theory of polyclonal activation produced interesting results, similar to those obtained by residual antigen theory, but without requiring a long-lived supply of antigen. The signalling provided by IL-15 seems to be essential for this phenomenon. It appears consistent with nature's efficient ways that the body would use the constant attack by antigen to strengthen itself, and we have demonstrated a polyclonal memory effect that is qualitatively similar to the experimental observations of [Bernasconi *et al.* 2002].

## 5.5 Further Work

The logical extension of our basic model of polyclonal memory is to create a more detailed B-cell/T-cell and APC model, and then to use that as the basis for a combined model attempting to simulate the latest theories of both B- and T-cell memory. Once such a model has been implemented, we can begin to explore questions specifically surrounding the relationship between B- and T-lymphocyte memory, and look at new rules for plasma cell and memory cell creation, death and homeostasis.

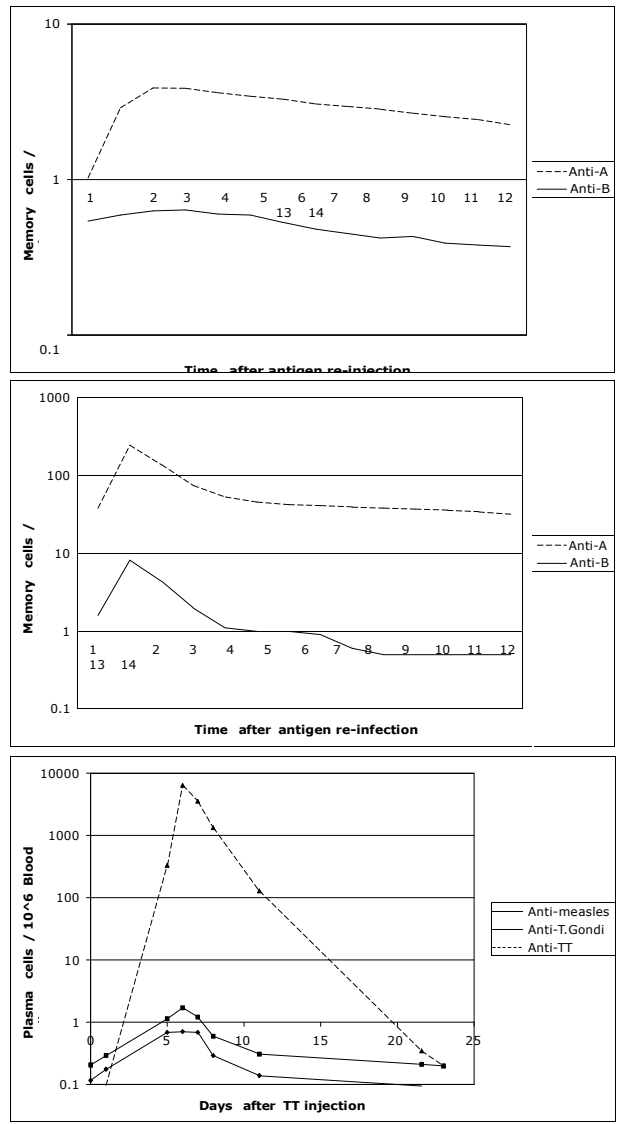

Fig. 5.9. (top and middle) Plots of the memory cell-levels per volume for two antigens, A and B, which are too dissimilar to directly cause a response in each other's memory cells. The immune system has already been exposed to both Antigen A and B; Antigen A is re-introduced at t=0. (top) and (middle) are for two different model parameterisations (see text). Both cases show an unexpected increase in the memory cells that are specific to the non-injected antigen. Since plasma cells levels are roughly linear, relative to memory cell levels, the *in silico* results are qualitatively consistent with the *in vivo* Bernasconi's results (bottom).

As mentioned above, the level of detail of a simulation should be as simple as possible, but a simulation that is too simple will not be as effective. This is a standard dilemma of machine learning hypothesis generation, and we intend to address this issue by means of automatic feedback. In other words, we will generate a *population* of simulations, and then evolve them to find the simplest, most effective candidate model.

The choice of parameters for any model is known to be a hard problem [Ljung 1999], but creation of the model is much harder [King *et al.* 2005]. We are examining several methods of assisted parameterisation of the models, so that a 'best-fit' can be found by Sentinel. This will allow the research to focus on the scientifically interesting model-building task, rather than the more mechanical parameterisation task, and will help to remove four of the possibilities for the differences in between the *in silico* and *in vivo* results in the previous section.

One of our long-term goals is to produce an integrated model of immunological memory that explains the experimental evidence used to support many of, if not all, the theories explored here. Such a model could be used to explore more detailed issues in immunological memory, such as the unusual effects of the SAP gene (which controls long-term memory, but has no effect on short-term memory) [Crotty *et al.* 2003]. Furthermore, a general theory of immunological memory would have implications for machine learning.

The applications described here are mostly related to immunology, and indeed that is the main focus of our work. Nonetheless, our Sentinel platform is likely to be useful in AIS endeavours in the future, in particular when it comes to understanding the dynamics of AIS algorithms that are based on complex systems of agents. In addition, simulating theories from immunology that have yet to be adapted by AIS researchers can provide assistance in determining the minimum set of features required in developing an abstract representation of an immune mechanism.

As Sentinel continues to develop, and becomes ever more sophisticated, we will be able to develop larger, more complex models than at present. It will be interesting to see if the increase in complexity is important, or whether there is a level of complexity that is sufficient for the majority of immunological research.